%% file: main.tex
\documentclass[final]{CVPR2021/cvpr}

\usepackage{times}
\usepackage{epsfig}
\usepackage{graphicx}
\usepackage{amsmath}
\usepackage{amssymb}

\usepackage{soul}
\usepackage[numbers,sort,compress]{natbib}

\usepackage{multirow}

\usepackage{makecell}
\usepackage{float}

\usepackage[pagebackref=true,breaklinks=true,colorlinks,bookmarks=false]{hyperref}

\usepackage{cleveref}

\input{CVPR2021/macros}

\title{SWAGAN: A Style-based WAvelet-driven Generative Model}

\author{Rinon Gal\\
Tel-Aviv University
\and
Dana Cohen\\
Tel-Aviv University
\and
Amit Bermano\\
Tel-Aviv University
\and
Daniel Cohen-Or\\
Tel-Aviv University
}

\begin{document}

\maketitle
\input{CVPR2021/abstract.tex}

\input{CVPR2021/intro.tex}

\input{CVPR2021/related.tex}

\input{CVPR2021/method.tex}

\input{CVPR2021/experiments.tex}

\input{CVPR2021/future.tex}

\small
\bibliographystyle{plainnat}
\bibliography{main}

\end{document}

%% file: CVPR2021/macros.tex
\usepackage{comment}
\usepackage{multirow,bigdelim}

\usepackage{array}
\usepackage{wrapfig}

\usepackage{adjustbox}
\usepackage[percent]{overpic}
\usepackage{makecell}
\usepackage[normalem]{ulem}

\usepackage{blindtext}

\newcolumntype{C}[1]{>{\centering\let\newline\\\arraybackslash\hspace{0pt}}m{#1}}

\newif\ifdraft
\drafttrue

\ifdraft
\newcommand{\dcc}[1]{{\color{red}[\textbf{DC:} #1]}}
\newcommand{\rgc}[1]{{\color{purple}[\textbf{RG:} #1]}}
\newcommand{\dac}[1]{{\color{olive}[\textbf{DC4:} #1]}}
\newcommand{\abc}[1]{{\color{blue}[\textbf{AB:} #1]}}

\else
\newcommand{\dcc}[1]{}
\newcommand{\rgc}[1]{}
\newcommand{\abc}[1]{}
\newcommand{\dac}[1]{}

\fi

\usepackage[bottom]{footmisc}

%% file: CVPR2021/abstract.tex
\begin{abstract}

In recent years, considerable progress has been made in the visual quality of Generative Adversarial Networks (GANs). Even so, these networks still suffer from degradation in quality for high-frequency content, stemming from a spectrally biased architecture, and similarly unfavorable loss functions. To address this issue, we present a novel general-purpose Style and WAvelet based GAN (SWAGAN) that implements progressive generation in the frequency domain. SWAGAN incorporates wavelets throughout its generator and discriminator architectures, enforcing a frequency-aware latent representation at every step of the way. This approach yields enhancements in the visual quality of the generated images, and considerably increases computational performance. We demonstrate the advantage of our method by integrating it into the SyleGAN2 framework, and verifying that content generation in the wavelet domain leads to higher quality images with more realistic high-frequency content. 
Furthermore, we verify that our model's latent space retains the qualities that allow StyleGAN to serve as a basis for a multitude of editing tasks, and show that our frequency-aware approach also induces improved downstream visual quality.

\end{abstract}

%% file: CVPR2021/intro.tex
\section{Introduction}
\label{sec:intro}

Image synthesis is a cornerstone of modern deep learning research, owing to the applicability of deep generative networks to a multitude of image related tasks. Such tasks range from content generation to novel view synthesis, rendering, image in-painting, super resolution or modifications of specific image properties \citep{thies2019deferred,ledig2017photo,karras_style-based_2019,nguyen2019hologan}. The performance of such downstream tasks has seen rapid improvements in recent years, due in part to the significant improvement in the visual quality of generative models, and in particular Generative Adversarial Networks (GANs).

\begin{figure}[t]
\begin{center}
\includegraphics[width=\linewidth]{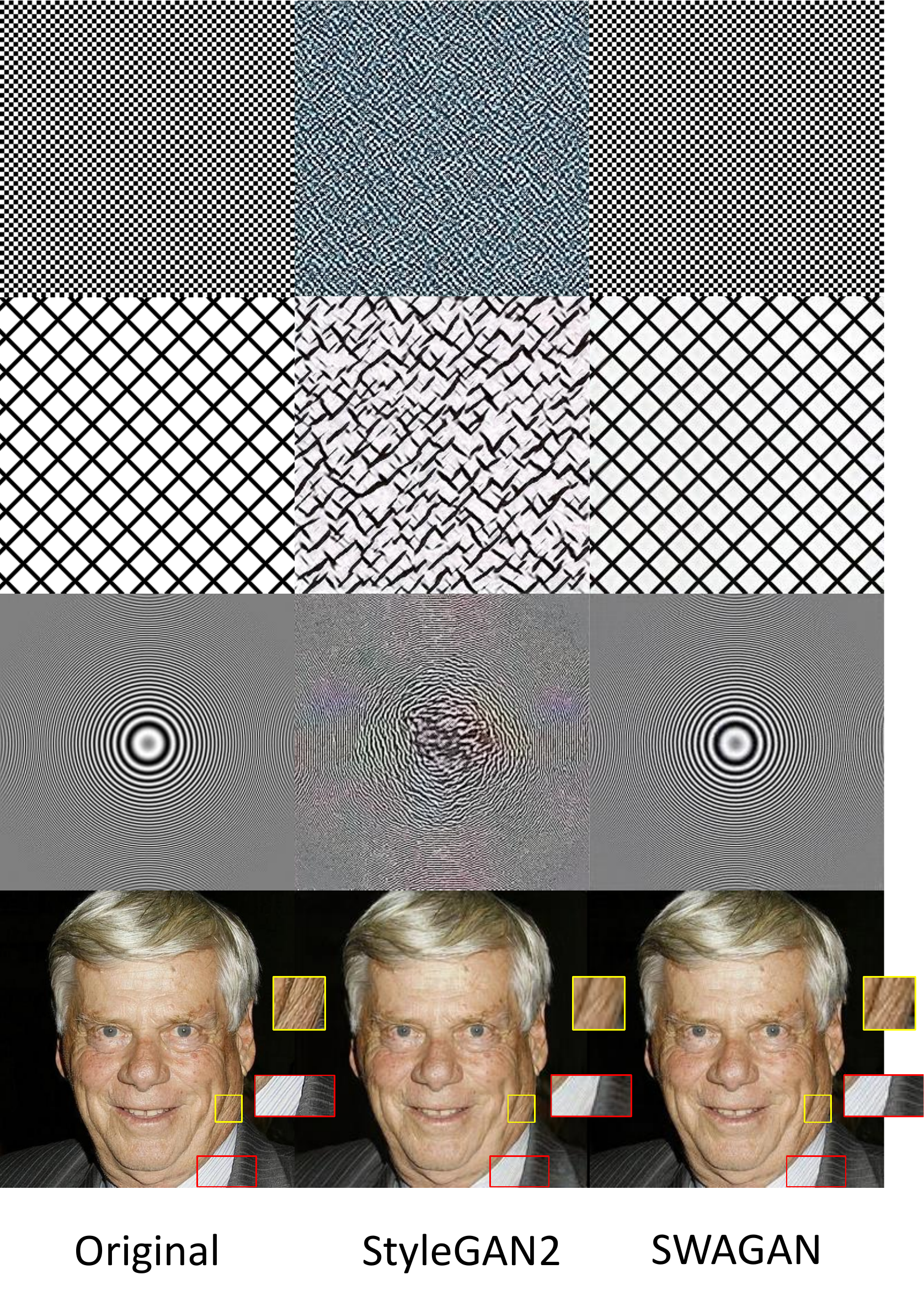}
\end{center}
   \caption{By working directly in the wavelet-domain, rather than image or feature space, we can alleviate the spectral bias of neural networks and successfully generate high frequency data where other models fail. Our model can produce patterns that elude state-of-the-art models such as StyleGAN2 even in an over-fitting setup, where the training set includes only a single image. We show (from left to right): the original image and the outputs of StyleGAN2 and SWAGAN (our model), respectively, after being trained for 24 hours on each image. Zooming in is encouraged.
   }
\label{fig:overfit}
\end{figure}

However, both generative models and their associated downstream tasks consistently suffer from a common flaw - their performance degrades rapidly when handling high frequency content \citep{durall2020watch,dzanic2019fourier,chen2020ssd}. This demeaned performance often manifests as blurriness, or the lack of sharp edges around finer image content. Prior works have attributed these shortcomings to the nature of the network architecture and common cost functions. The oft-used mean squared reconstruction error, for example, is much less sensitive to small perturbations, thus leading networks that utilize this loss to favor the accuracy of low frequency and large-scale content over finer features \citep{huang2017wavelet}. In addition, recent works have pointed to an inherent spectral bias in neural networks that manifests as a frequency-dependent learning speed \citep{rahaman2019spectral}.

A straightforward approach to combat this shortcoming is to push resolutions to new heights. By doing so, previously minute details become more dominant and thus are better captured by the traditional losses. This approach, however, does little to overcome the frequency-dependent learning rate. 

The current state-of-the-art results for high resolution generation are achieved by style-based models \citep{karras_style-based_2019,karras2020analyzing}. These models employ a new approach to content generation - rather than progressively growing a set of features and converting them to an image representation at the very end, they first generate a low resolution image and then grow it in size in a hierarchical manner \citep{karras2017progressive,karnewar2019msg}. This approach affords additional regularization and stability to the trained model and has been shown to achieve improved visual results. While these models lead to remarkable strides in visual quality, they rely on \mbox{networks} which are larger and 
notoriously expensive to train, with the \mbox{development} of recent state-of-the-art networks requiring upwards of 50 years of computation time using a top of the line GPU \citep{karras2020analyzing}.

An alternative approach to addressing the high-frequency gap is to directly employ frequency-aware training objectives, for example by empowering a discriminator to consider not only an image, but also a spectral decomposition of it \citep{durall2020watch,dzanic2019fourier,chen2020ssd}. Others opt to generate high-frequency content directly by predicting a full set of wavelet coefficients. Such an approach has been used successfully on a range of tasks, from image super resolution \citep{huang2017wavelet, zhang2019image} to compressed video enhancement \citep{wang2020multi}. However, to the best of our knowledge, these prior works focus on training a constrained generator for a specific task (e.g., face super resolution), and none of them are adapted to the modern approach of hierarchical image growth, rather than purely feature-based growth. 

We argue that frequency-based representations hold another advantage that is often overlooked or squandered due to misuse. By using an appropriate representation, the network can be biased to affect high-frequency modifications in some target domain through low frequency modifications of the representation. Such bias can help bypass the need to learn any high-frequency functions and thus avoid the spectral bias of neural networks.

In our work, we propose to marry the core ideas from these different approaches. Our goal is to train a general purpose image generator. However, rather than creating an image by predicting deep features or the color content of each pixel in a progressively growing series of resolutions, we instead propose to work directly in frequency space by predicting coefficients of basis functions of increasing \mbox{frequency}. Similarly, we propose to make use of a \mbox{discriminator} that compares not only the pixel content of generated images and real samples, but also their frequency space decomposition. These frequency domain \mbox{considerations} are incorporated not only as a training goal, but are baked directly into the architecture of our model at each resolution step. 

\begin{figure}[t]
\begin{center}
\includegraphics[width=\linewidth]{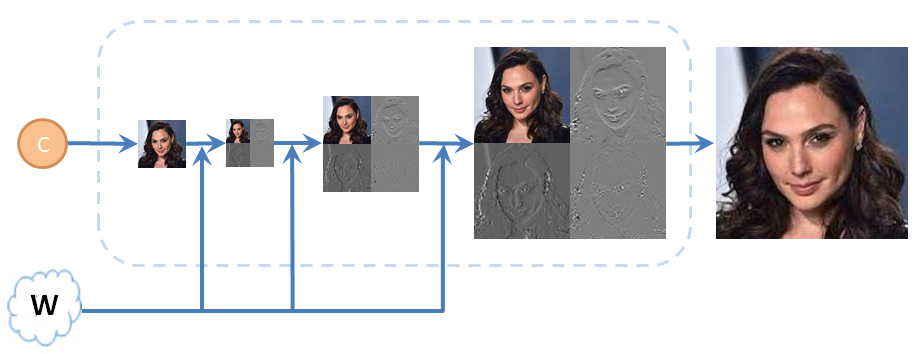}
\end{center}
   \caption{ Our style-based generator architecture predicts wavelet-coefficients at increasing resolution scales. 
   }
\label{fig:Architecture}
\vspace{6pt}
\end{figure}

We demonstrate the benefits of our approach using the popular StyleGAN2 framework \citep{karras2020analyzing} (see Figure~\ref{fig:Architecture}). We show that by adapting its architecture to work directly in a wavelet-based space, we achieve improved visual quality and more realistic content in the high-frequency range. By working directly in the wavelet-domain, rather than \mbox{image} or feature space, we alleviate the spectral bias of \mbox{neural} \mbox{networks}, converge with fewer epochs, and successfully generate high frequency data where other models fail (see Figure~\ref{fig:overfit})

Another surprising benefit of our approach is a significant reduction in training iteration and inference times. Our experiments show that the wavelet-based model not only converges with fewer epochs, but also that the tradeoff between spatial dimensions and channel depth inherent to wavelet decomposition, results in a significant speedup of network computations. This behavior can be dependent on the specific details of the convolution operations implemented by each framework, but popular implementations such as GEMM or Winograd variants show only a mild \mbox{increase} in computation time when increasing the number of filters \citep{jorda2019performance}.

Finally, we show that our model's latent space can support the editing operations that are commonplace in StyleGAN-based works. We show that our high-frequency performance can be successfully transmitted downstream, and that by using our model one can avoid many of the blurry artifacts present in prior works. Our experiments \mbox{indicate} that shifting content generation to the wavelet \mbox{domain} offers a significant set of advantages, with no \mbox{apparent} drawbacks.
\\
In summary, our key contributions are:
\begin{itemize}
    \item A hierarchical, wavelet-based approach to image generation that achieves improved visual fidelity and more realistic \mbox{spectra}.
    \item An approach to network design that reduces the natural spectral-bias inclination of networks, resulting in a significant reduction of the computational budget required to train high resolution generative models.
\end{itemize}

Our code and trained models will be made public.

%% file: CVPR2021/related.tex
\section{Background and related work}
\label{sec:related}

\subsection{Generative Adversarial Networks}

Generative Adversarial Networks (GANs), first proposed by \citet{goodfellow_generative_2014}, have been successfully used in a wide range of image synthesis tasks, ranging from general content creation to more specific tasks such as image super resolution and in-painting. Due to their ubiquity, considerable effort has gone into improving the quality of images generated by such models. Earlier models \citep{radford2015unsupervised,chen2016infogan,mirza2014conditional} performed content generation in feature space, growing a feature representation in a hierarchical manner before converting the final feature set into an image.

More recently, progressive \citep{karras2017progressive} and style-based \citep{karras_style-based_2019,karras2020analyzing} generative networks have appeared. These models achieved unprecedented resolutions and visual fidelity and solidified themselves as leading paradigms in the content generation field. One of the core differences between these models and previous generative works is that they generate hierarchical content directly in image space, without resorting to a conversion from learned features at the very end. This is achieved through direct supervision in the progressive case, or merely by applying naive bilinear-upsampling between different scales and teaching the network to augment them with high-resolution detail as in \citet{karras2020analyzing}. These additional constraints serve to regularize the network, preventing any learned features from deviating too widely and eventually improving the quality of results.

Our work aims to improve upon this idea by progressively generating image decompositions in a frequency domain. Rather than training the different layers to implicitly add higher-frequency data to an image by modifying its color content, we train them to do so explicitly by modifying the coefficients of a wavelet band decomposition. Similarly, our discriminator extracts features not only from the image space representations but also from wavelet domain sub-bands. The network is thus empowered to better identify content that lacks high-frequency data, and can more directly restore it.

\subsection{Frequency-based Approaches}

The use of frequency based information, and in particular wavelet transforms, has been well studied in the deep learning literature. Convolutional neural networks have been augmented with wavelet-based features, inputs or pooling operations \citep{gao2016hybrid, liu2019multi, williams2018wavelet} in order to improve performance on style transfer \citep{yoo2019photorealistic}, denoising \citep{liu2020wavelet} and even medical analysis tasks \citep{kang2017deep}. Others \citep{tancik2020fourier,mildenhall2020nerf} have proposed a decomposition of implicit representation input coordinates into a set of Fourier-based features. They showed, both empirically and theoretically, that explicitly providing the network with high frequency inputs allows it to overcome spectral bias and better reconstruct high-frequency outputs. 
In the generative domain, several works \citep{huang2019wavelet, wang2020multi, zhang2019super} have proposed empowering networks to create not only images, but also full wavelet based decompositions. Others \citep{liu2019attribute, wang2020multi} have augmented discriminators by providing them with not only an image, but also frequency sub-bands derived from a wavelet packet transform.

None of these works, however, are general purpose generative models. Instead, they utilize adversarial losses in order to improve performance on specific tasks such as image super resolution or denoising. As such, none of them can generate a novel image from a latent code but rather only add or improve upon existing images.

Our work, meanwhile, seeks to train a fully generative model which can synthesize novel images from noise samples. Such a generator can later be adapted to a multitude of tasks by projecting images into its richer latent space, either through direct optimization \cite{abdal2019image2stylegan} or through the use of a trained encoder \citep{richardson2020encoding}. This methodology has shown considerable success since the inception of powerful generative models such as \citet{karras_style-based_2019}.

More recently, a significant number of studies \citep{chen2020ssd, durall2020watch} have highlighted the failures of general-purpose generative \mbox{networks} in the high-frequency domain. All of them, however, address the problem by introducing frequency related terms to the loss function. In contrast, we modify not only the loss term, but also introduce a frequency-based inductive bias into the generator itself. This bias is shown to further bolster the visual quality of synthesized images, while also leading to significant reductions in training and \mbox{inference} times.

Lastly, our work is further distinguished from prior wavelet-based approaches in that we adopt the modern viewpoint of progressive generation in the frequency domain. Whereas previous works opted to utilize a progressive feature representation and generate wavelet content only at the final layer, or alternatively provide the discriminator with wavelets at the onset but merely encode them into features - our model generates content directly in wavelet space throughout the entire process.

%% file: CVPR2021/method.tex
\section{Method}
\label{sec:method}

\subsection{Wavelet Transform}
At the core of our method are wavelet transforms, which decompose an image into a sequence of channels, each of which represents a different range of frequency content.
We employ Haar Wavelets as the basis function for our transformations, owing to their simple nature coupled with their well documented ability to represent multi-frequency information \citep{daubechies1990wavelet,daubechies1992ten}. Our model works with the first-level wavelet decomposition, where each image is broken into four sub-bands: LL, LH, HL and HH. These bands are attained by passing the image through a series of low-pass and high-pass wavelet filters and serve as the wavelet coefficients. The first sub-band, LL, corresponds to low frequency information, and indeed it is visually similar to a blurred version of the input image. The remaining sub-bands, LH, HL and HH, correspond to high frequency content in the horizontal, vertical and diagonal directions, respectively.

\subsection{Network Architecture}

\begin{figure*}
\begin{center}
\includegraphics[height=6cm]{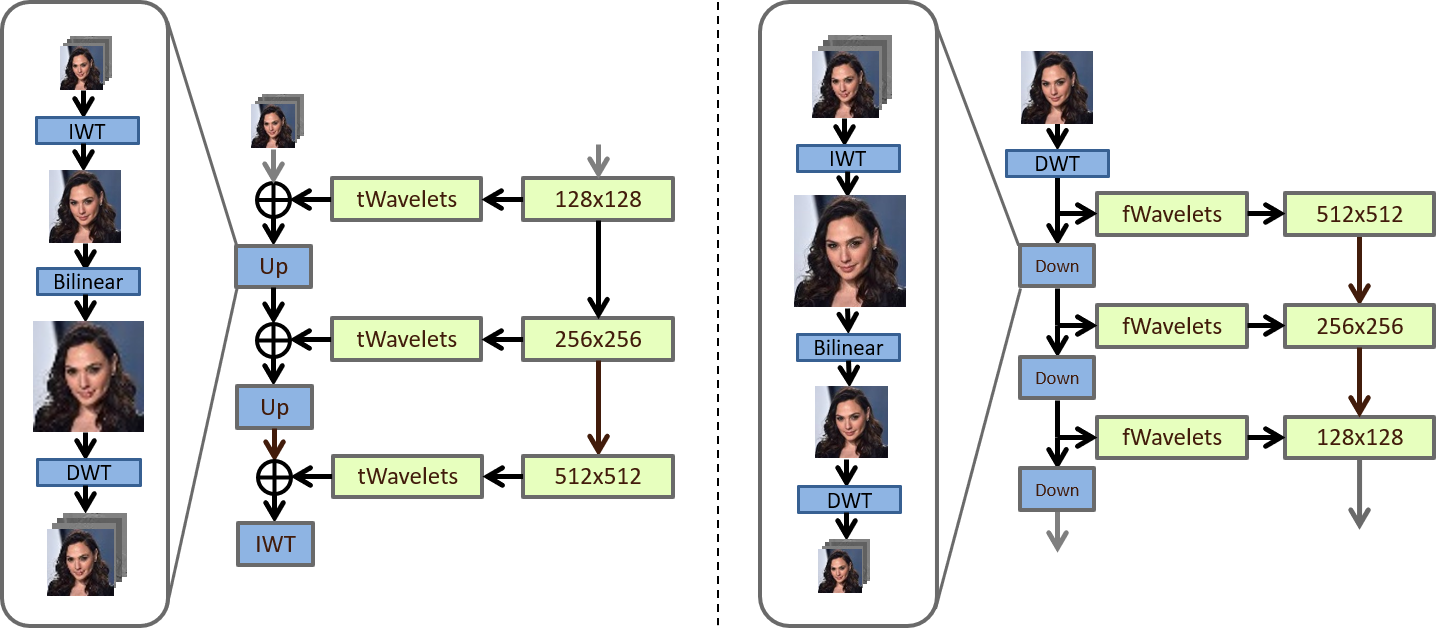}
\end{center}
   \caption{Our SWAGAN generator (left) and discriminator (right) architectures. Each ConvBlock is equivalent to a feature-resolution increasing block of the StyleGAN2 architecture, which is itself composed of two style blocks. \textit{tWavelets} and \textit{fWavelets} correspond to the tRGB and fRGB layers of StyleGAN2 and their purpose is to learn a mapping between wavelet decompositions and high dimensional features. Inverse wavelet transforms are denoted by \textit{IWT}, while \textit{Up} and \textit{Down} are non-learning layers responsible for converting an image to an initial wavelet-decomposition of a higher or lower resolution, respectively.}
\label{fig:Architecture}
\end{figure*}

Our wavelet-aware architecture is based on the \mbox{original} implementation of StyleGAN2 \citep{karras2020analyzing}. While the original implementation generates content directly in image-space, our proposed architecture operates in the frequency domain. Similarly, our discriminator is designed to consider not only the RGB space of images, but their entire wavelet \mbox{decomposition}. 

By allowing the generator to generate content directly in the wavelet domain, we gain on two fronts: First, as shown by \citep{rahaman2019spectral}, neural networks prioritize learning in the low frequency domain. By converting our representation to a frequency based one, we empower the network to affect \textit{high-frequency} changes in the \textit{image domain} by learning \textit{low-frequency} modifications of the \textit{representation}. This differs from a simple loss-based modification, as the later can provide an incentive for the network to learn high-frequencies, but it does not make the learning task easier. 
Second, wavelet decompositions are more spatially compact. In a first-level wavelet decomposition, each $2N\times 2N$ image is fully represented by four channels of $N\times N$ coefficients each. This allows us to employ convolutions on lower resolution representations throughout the entire generative process, at the cost of requiring additional filters. However, this tradeoff can be advantageous when utilizing the popular deep learning frameworks \citep{jorda2019performance}.

Similarly, by providing frequency information to the discriminator, the network is able to better identify the high frequency content that is often missing in generated images. As a result, the generator is better motivated to re-create plausible high-frequency data. 

In the proposed generator, each resolution block receives a full wavelet decomposition as an input. In a similar fashion to StyleGAN2, the wavelet coefficients are refined using a set of high dimensional features mapped back to the frequency domain via skip connections. In StyleGAN2, the image is resized between blocks using simple \mbox{bilinear} \mbox{up-sampling}. Performing this up-sampling in the \mbox{frequency} domain would not carry the same meaning. Instead, we choose the natural alternative and perform up-sampling by converting the wavelet representation back to the image domain by applying the inverse wavelet transform (IWT), \mbox{resizing} the image as usual, and then predicting the next set of wavelet coefficients from the higher-resolution \mbox{image}. The network's output is formed by applying an IWT to the wavelet decomposition provided by the output of the last layer. The generator's architecture is illustrated in Figure~\ref{fig:Architecture} (left).

In the discriminator, we similarly provide the wavelet coefficients of the corresponding resolution to each block. At each resolution step, a skip-connection based network is used to extract features from the wavelet decomposition and merge them into the feature representations derived from higher resolution blocks. In order to downscale the image between blocks, the wavelet coefficients are combined back into a full image via IWT, the image is bilinearly down-sampled, and the lower-resolution image is decomposed back into the wavelet coefficients, which are fed into the next block. The input to the first block of the discriminator is simply the DWT of any image (real or fake). The discriminator's architecture is illustrated in Figure~\ref{fig:Architecture} (right).

In addition to the architecture described above, we explored a series of variations, including different up-sampling steps and mixing image and wavelet-domain generation steps in the same network. We describe and analyze these variations in Section \ref{sec:exp}.

\subsection{Downstream Tasks} \label{sec:inversion}
Any adaptation of the style-based network would be incomplete if it could not be used as a backbone for the downstream tasks built around the original. We demonstrate that our wavelet-based network can still support the same applications, and in some cases even achieve better results.

We analyze our model using the optimization-based approach to inversion, where we use gradient descent to find the latent space representation for which the generator outputs the best approximation of a given source image. We use the latent space projector of \citet{karras2020analyzing}, which finds a latent representation in the so-called W space of a style-based network, using a target of minimizing a perceptual based loss (LPIPS). The later is crucial for our needs, as any L2 based reconstruction target will inevitably discard high-frequency information in the optimization process.

\subsection{Implementation and Training Details}
We build upon the official TensorFlow implementation of StyleGAN2 \citep{karras2020analyzing}. Specifically, we modify the architecture of the skip-based generator and discriminators, but otherwise inherit all of the corresponding training details and parameters from the configuration F setup. Our models were trained on FFHQ at resolutions of $256\times 256$ and $1024\times 1024$ and on LSUN Churches at a resolution of $256\times 256$. For FFHQ at $1024\times 1024$, we utilized a batch size of 8 for the wavelet-based models and 2 for StyleGAN2 comparisons. For FFHQ at $256\times 256$ resolution, we used batch sizes of 16 and 8, respectively. All models were trained on a single NVIDIA GeForce RTX 2080 GPU, i.e., using commercially available, non-corporate grade computational power.
For quantitative evaluations of downstream tasks, we utilized the publicly available CelebA dataset \cite{liu2018large}.

%% file: CVPR2021/experiments.tex
\section{Experiments}
\label{sec:exp}

We have conducted a series of experiments in order to evaluate the increased computational and visual performance of our generator, and to evaluate the importance of high frequency data to downstream tasks.
We additionally conduct an ablation study to highlight the importance of generation in the wavelet-domain, the importance of a wavelet-based discriminator and to investigate natural alternatives to some of our ad-hoc architectural choices.
In all of the experiments below, we dubbed the \textit{bilinear up-sampling} variation of our generator (presented in Section \ref{sec:method}) as SWAGAN-Bi.

\subsection{Computational Performance and Visual \mbox{Quality}}

We first demonstrate our ability to capture high-frequency data on a set of toy problems. We generated three datasets, each containing only a single image of a pattern with high-frequency detail. In addition, we created a fourth dataset composed of a single real face with considerable high-frequency data. We trained our model and the original StyleGAN2 model on these single image datasets for a period of 24 hours. Figure~\ref{fig:overfit} shows the outputs of the trained models. When dealing with repeated patterns, our model can cleanly capture and recreate them with high accuracy. The original StyleGAN2 model, meanwhile, fails to recreate the data in a meaningful manner, showing that purely high-frequency content is beyond its scope. The face image is well reconstructed by both models. However, our model is better able to capture details such as the patterns on the shirt or fine wrinkles.

\begin{figure*}[!hbt]
\begin{center}
\includegraphics[width=\linewidth]{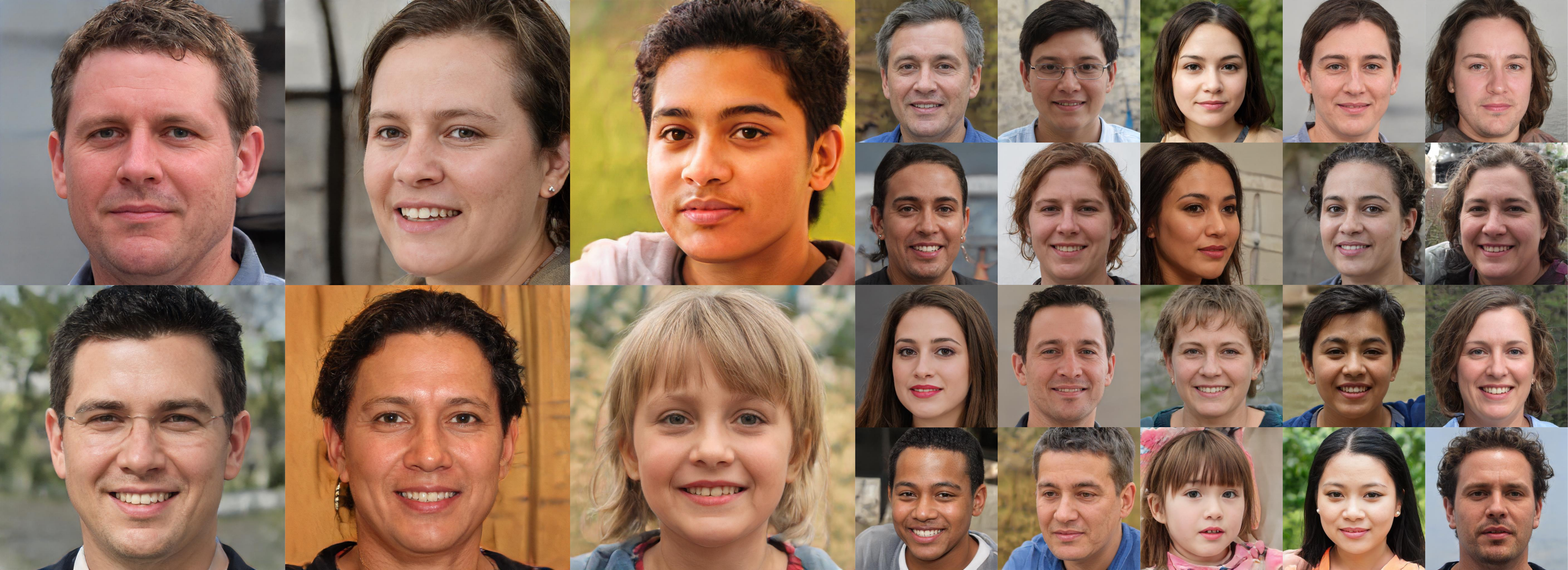}
\end{center}
   \caption{Uncurated face samples generated by our SWAGAN-Bi model, using truncation with $\psi = 0.5$.}
\label{fig:fig_gen_1024}
\end{figure*}

\begin{figure}[!hbt]
\begin{center}
\includegraphics[width=\linewidth]{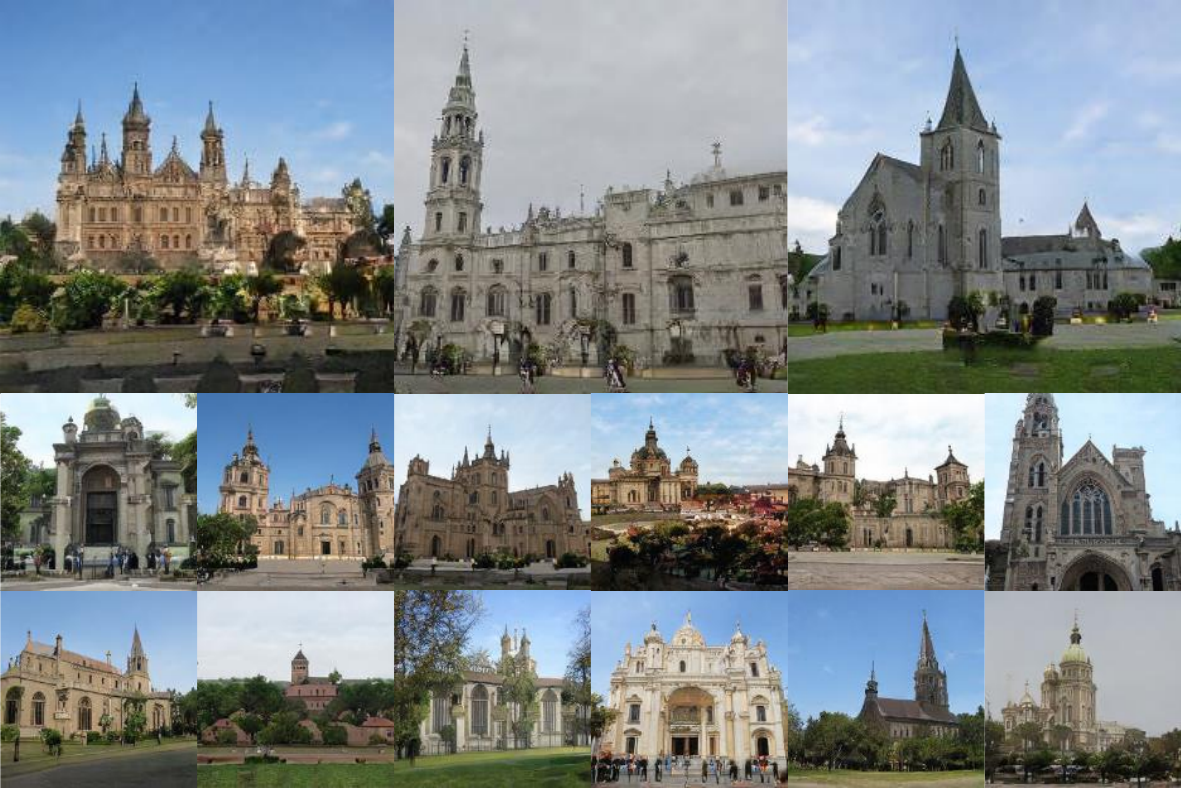}
\end{center}
   \caption{Uncurated samples generated by our SWAGAN-Bi model trained on the LSUN Church dataset at a resolution of 256x256, using truncation with $\psi = 0.5$.}
\label{fig:fig_gen_256}
\end{figure}

We compared the time required to train our model and the visual quality of generated results throughout the process to those achieved by StyleGAN2. Figure~\ref{fig:fig_gen_1024} shows an uncurated sample of faces drawn from SWAGAN-Bi model trained on FFHQ at 1024x1024.
Figure~\ref{fig:fig_gen_256} shows uncurated samples drawn from a SWAGAN-Bi model trained on LSUN Church at a resolution of 256x256. As can be seen, our model can produce high quality results with significant detail around regions of high frequency-content, such as wrinkles or hair.

In Table \ref{tab:train_times}, we report the time required for the model to process 1000 real images on a single GeForce RTX 2080. This metric is equivalent to time per epoch (up to a \mbox{constant}, multiplicative factor), and is the standard unit of work used in \cite{karras_style-based_2019} and its derivatives. The wavelet-based variants require a significantly reduced training time across both \mbox{resolutions}. 

Table \ref{tab:fid_compare} reports the final Fréchet Inception Distance (FID) of our models at various resolutions and on multiple datasets. As the computational overhead for training a model for the number of iterations reported by StyleGAN2 remains high, we elected to reduce the resolution and the number of training steps used for training the models. In all such cases, we provide results on a StyleGAN2 model trained for the same number of iterations and resolution, utilizing the official implementation of \citet{karras2020analyzing}. Note that this comparison favors the StyleGAN2 models by allotting them roughly double the computational budget, due to increased training times and higher memory requirements. Even under this biased comparison, the wavelet-based approach can capture more realistic details and achieve better visual quality.

\begin{figure*}[t]
\begin{center}
\includegraphics[width=\linewidth]{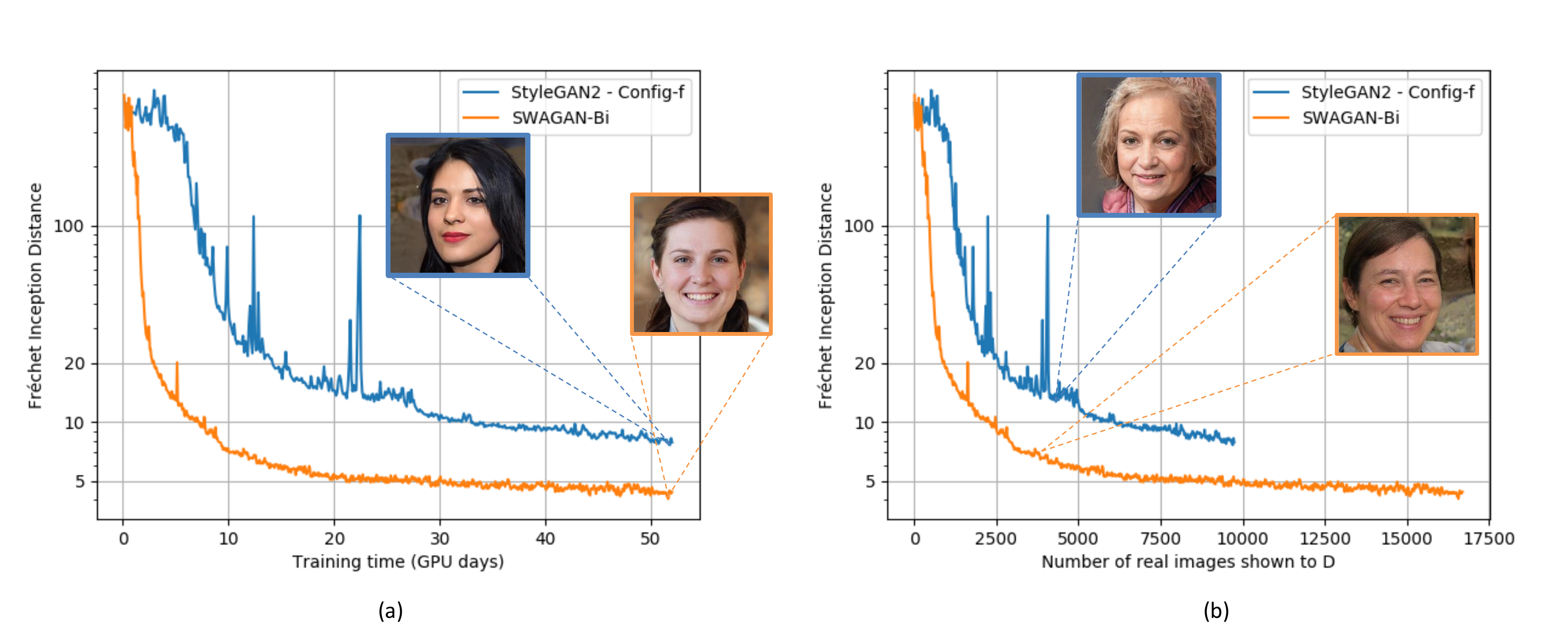}
\end{center}
   \caption{Model FID as a function of: (a) wall-clock time, (b) number of images viewed by the discriminator. Our model outperforms StyleGAN2 in both scenarios and displays increased stability during training.}
\label{fig:fid_compare_1024}
\end{figure*}

Figure~\ref{fig:fid_compare_1024} shows a comparison of FID metrics over the duration of training between a wavelet based model (SWAGAN-Bi) and the original StyleGAN2 model, trained on FFHQ at 1024x1024 resolution. We show the progress, both as a function of the number of images viewed by the discriminator and as a function of wall-time. Our model, not only trains faster, but also converges with fewer epochs on the same data set. This hints that the inductive bias built into our model is indeed capable of alleviating the spectral bias of neural network learning. 

\begin{table}[!hbt] \footnotesize
    \centering
    \begin{tabular}{lcc}\hline
        \bf{Model} & \bf{Resolution} & \bf{Seconds / 1k images $\downarrow$} \\\hline
        StyleGAN2 & 1024 & 184.97 \\
        SWAGAN-Bi & 1024 & 95.06 \\ \hline
        StyleGAN2 & 256 & 143.04 \\
        SWAGAN-Bi & 256 & 80.07 \\
        SWAGAN-Bi w/ NWD & 256 & 129.65 \\
        SWAGAN-NU & 256 & 70.06 \\
        Wavelet at final layer & 256 & 116.06 \\ \hline
    \end{tabular}\\
    \caption{Average time (in seconds) for the network to process 1000 real images during training. Wavelet-based generators consistently outperform the alternatives. The reported times are based on the calculation provided in StyleGAN2's original implementation and do not include any of the (significant) time spent on benchmark calculations or sample generation.}
    \label{tab:train_times}
\end{table}

\begin{table}[!hbt] \footnotesize
    \centering
    \begin{tabular}{lccc}\hline
        \bf{Model} & \bf{Dataset} & \bf{Images Viewed} & \bf{FID $\downarrow$} \\\hline
        StyleGAN2 & FFHQ 1024 & 9.76M & 7.62 \\
        SWAGAN-Bi & FFHQ 1024 & 9.76M & 4.68 \\
        SWAGAN-Bi & FFHQ 1024 & 17.1M & 4.06 \\ \hline
        StyleGAN2 & FFHQ 256 & 8.5M & 5.82 \\
        SWAGAN-Bi w/ NWD & FFHQ 256 & 8.5M & 12.27 \\
        SWAGAN-Bi & FFHQ 256 & 8.5M & \bf{5.22} \\
        Wavelet at final layer & FFHQ 256 & 8.5M & 7.96 \\ 
        SWAGAN-NU & FFHQ 256 & 8.5M & 9.20 \\ \hline
        StyleGAN2 & Church 256 & 6.5M & 5.46 \\
        SWAGAN-Bi & Church 256 & 6.5M & \bf{4.97} \\
        \hline
    \end{tabular}\\
    \caption{The best FID achieved by all tested setups until the discriminator observed the indicated number of images. Our SWAGAN-Bi model achieves the best results for different resolutions and datasets.}
    \label{tab:fid_compare}
\end{table}

\begin{figure}[t]
\begin{center}
\includegraphics[width=\linewidth]{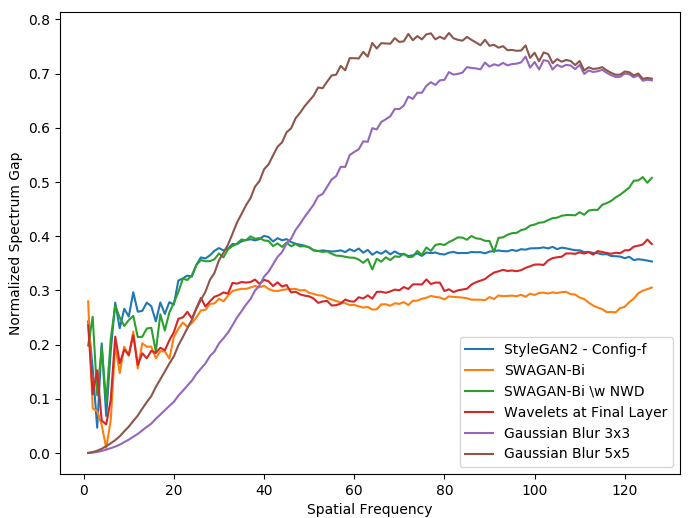}
\end{center}
   \caption{The power spectrum distance between real images and those produced by the listed models, normalized by the spectra of real images. To assist comparisons, we also provide the spectrum gap observed in real images blurred using Gaussian kernels of size 3x3 and 5x5. Our full model (orange) achieves more realistic results in the middle and high frequency range. Note that a model without a wavelet-based discriminator, or one where the generative process is not entirely wavelet, based both fail to capture the same degree of high frequency data. }
\label{fig:spectrum_256}
\end{figure}

In Figure~\ref{fig:spectrum_256}, we plot the distance between the power spectrum of real images and those generated by a subset of our trained models, averaged over 4050 images per model and normalized by the spectra of real images (dubbed "spectrum gap"). For convenience, we additionally plot the spectrum gap for real images blurred via Gaussian kernels with filter sizes of $3\times 3$ and $5\times 5$. Our frequency-domain model consistently generates more realistic spectra, particularly in the medium and high frequency range.

\subsection{Ablation Study} 

In order to measure the importance of each aspect of our suggested solution, we asked ourselves the following questions, and conducted a series of experiments to answer each in turn. The quantitative results of these experiments are provided in \cref{tab:train_times,tab:fid_compare} and in Figure~\ref{fig:spectrum_256}.

\paragraph{\bf{Is it important to progressively generate content in the frequency domain?}}
We investigate this question with two experiments. In the first, we observe the behaviour of the network when all layers, but the last utilize the original tRGB mapping. In such a setup, content is progressively generated in the RGB domain and any wavelet information is only added at the final layer. The model which only predicts wavelet information in the final layer has shown inferior performance, both in time-based metrics and in visual quality, highlighting the importance of wavelet-domain generation throughout the entire process. Note that this model performs worse than even the original StyleGAN2. This may be due to the network's need to conserve both frequency and image based information in the hidden feature representation.

\begin{figure}[t]
\begin{center}
\includegraphics[height=4cm]{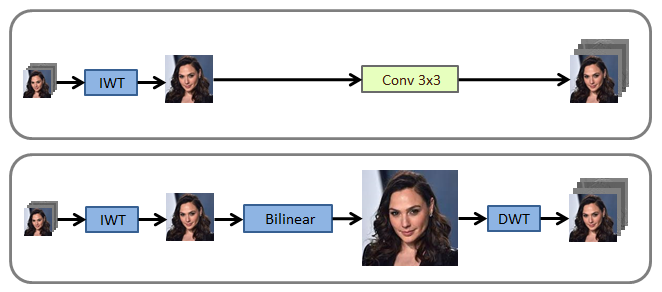}
\end{center}
   \caption{The upscaling blocks for SWAGAN-NU (top) and SWAGAN-Bi (bottom) architectures. SWAGAN-NU architecture includes $3\times 3$ convolutions in order to predict the new wavelet channels; The SWAGAN-Bi architecture is comprised of bilinear up sampling, followed by discrete wavelet transform (DWT).}
\label{fig:UpNet}
\end{figure}

Next, we investigated a variation on the model's up-scaling layer. In this setup, rather than resizing the image through bilinear up-sampling, we utilize a network to predict an initial set of wavelet coefficients for the next layer, effectively generating content at the next resolution step. This upscaling layer is visualized in Figure~\ref{fig:UpNet}. We dubbed this \textit{neural upscaling} variation as SWAGAN-NU. 

\begin{figure}[t]
\begin{center}
\includegraphics[width=\linewidth]{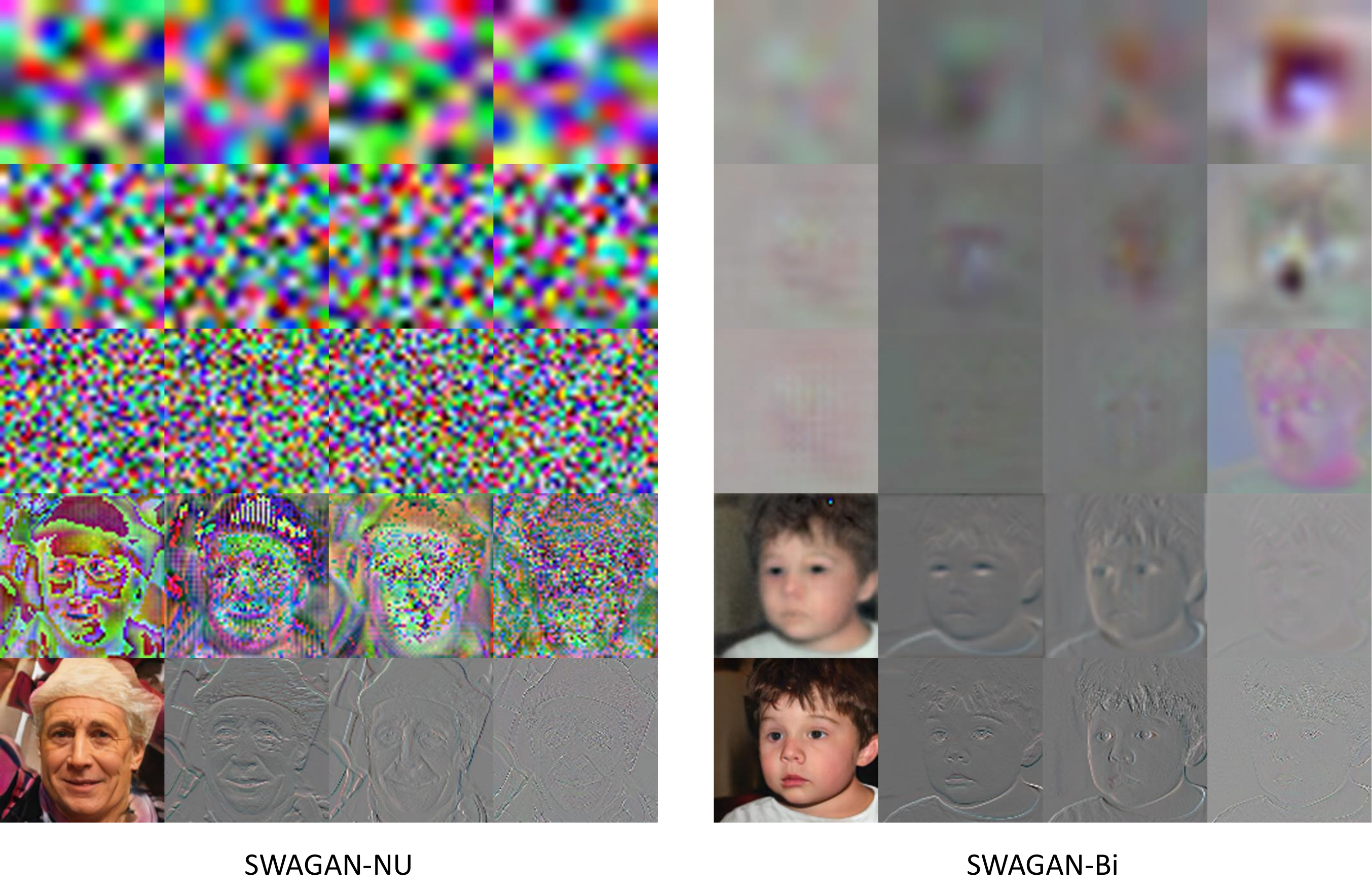}
\end{center}
   \caption{Intermediate wavelet representations in the generative path of our SWAGAN-Bi and SWAGAN-NU models. At each row, we plot the four wavelet coefficient maps obtained at a given resolution step, prior to the IWT operation. All outputs were resized to $256 \times 256$ using bicubic interpolation. The SWAGAN-Bi outputs portray an increasing level of detail in the wavelet space, while the SWAGAN-NU outputs portray a deep feature representation until the final resolution. Zooming in is encouraged }
\label{fig:layer_outputs}
\end{figure}

A reasonable expectation would be for the network to perform better in this scenario, as it has sufficient degrees of freedom to learn a better upsampling step. However, although it sounds promising, such a change also breaks the inductive bias built into the network. In the bilinear up-sampling approach, the network must truly generate wavelet based content at each scale because it has no means through which to convert the hidden representation back to the wavelet domain before the output layer. The added neural mapping of the SWAGAN-NU architecture, however, could also be used to learn a mapping from any feature representation back to wavelet space, thus allowing the network to ignore our intent and merely work in feature space throughout. The loss of this bias adversely affects the visual quality of generated results. We further examined this hypothesis by investigating the images observed in the intermediate scales of the network. A comparison of such images is provided in Figure~\ref{fig:layer_outputs}, where we plot the four wavelet maps predicted at increasing resolution scales, resized to $256\times 256$. As can be seen, intermediate representations in the SWAGAN-Bi generator portray a gradual growth in wavelet space, while those of the SWAGAN-NU setup do not.

As our ablation experiments show, progressive generation in the frequency domain is key to improving results.

\paragraph{\bf{Does the network benefit from modifying both the generator and the discriminator?}}
To answer this question, we trained a model utilizing the SWAGAN-Bi generator, along with the original residual based discriminator of StyleGAN2 (dubbed as NWD for non-wavelet-discriminator). The trained model offers significantly degraded performance when compared to the full SWAGAN-Bi model. This demeaned performance can be observed in both the FID metrics of Table \ref{tab:fid_compare} and in the power spectrum comparisons of Fig \ref{fig:spectrum_256}, where the average spectra of images generated by the full model more closely resemble that of real images. Indeed, the results of such a model are worse than those of the original StyleGAN2 across the board.
In addition to the degraded results, we also observe a significant increase in training instability when using a wavelet-based generator without a wavelet-based discriminator. When the loss term is insensitive to high-frequencies, the corresponding high-frequency bands in the wavelet decomposition receive diminished gradients, and the network fails to learn how to modify them in a meaningful manner. These results indicate that a wavelet-based generator should be accompanied by an appropriate frequency aware loss term (and the original StyleGAN2 discriminator is insufficient in that regard, as shown in \citet{chen2020ssd}). 

\section{GAN Inversion and Applications}

One of the key benefits of the style-based generative framework is its well behaved latent space, a property which can be exploited for editing tasks or for interpolating images in the generator's latent space. As we deem this to be a crucial property, we conducted a series of experiments to ensure that our latent space still supports these operations. 

\begin{figure*}[t]
\begin{center}
\includegraphics[width=\linewidth]{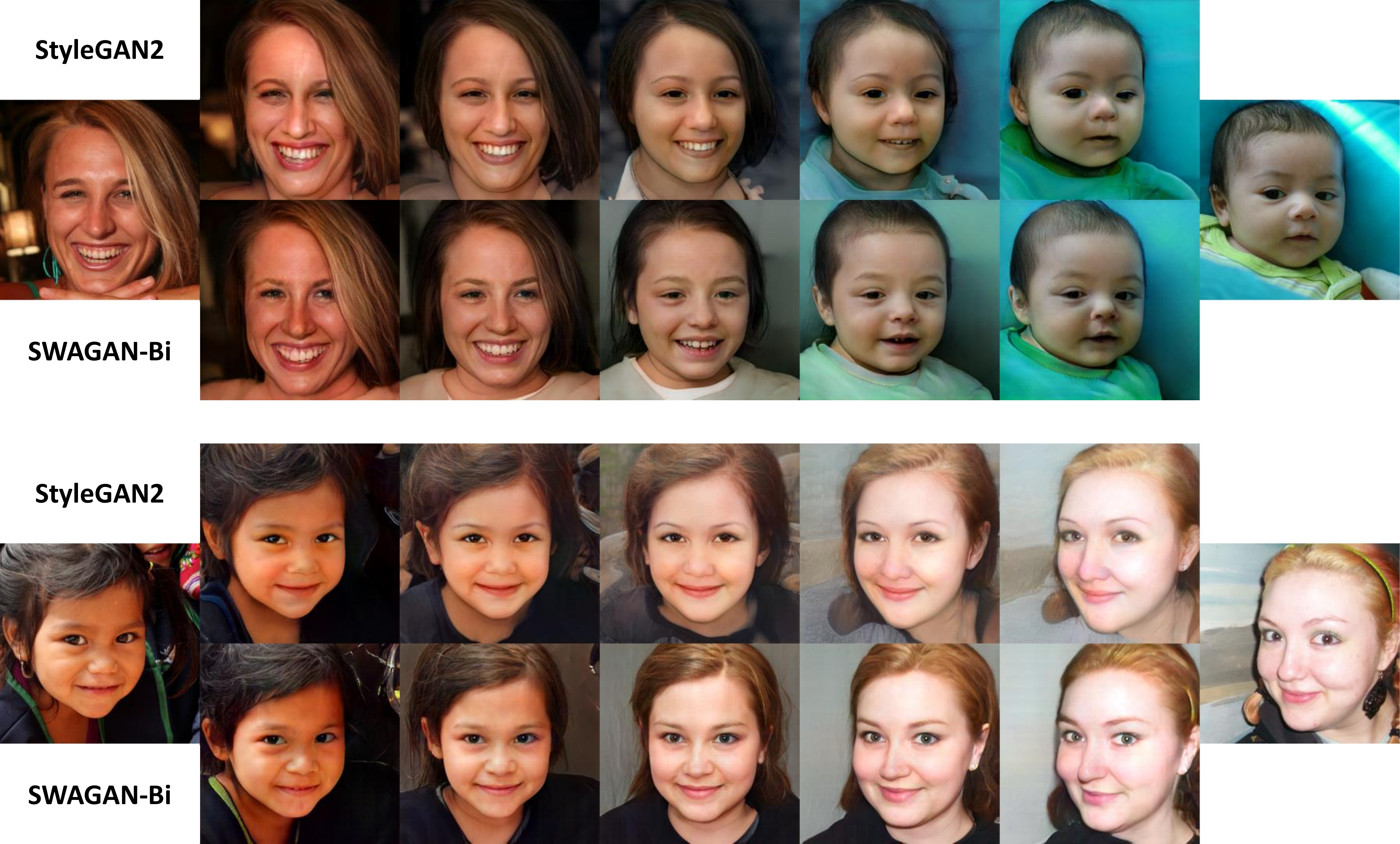}
\end{center}
   \caption{Latent space interpolation. In each cluster of images, the left-most and right-most images are the ground truth, taken from the FFHQ dataset. We project them into the generator's latent W-space using the StyleGAN projector, and interpolate between both ends. The top row of each cluster contains images obtained using a StyleGAN2 configuration F model, while the bottom row contains images obtained using the SWAGAN-Bi model. Interpolated StyleGAN2 images show considerable blur around regions of high frequency such as the hair, while the SWAGAN images do not.}
\label{fig:linear_1024}
\end{figure*}

\begin{figure*}[t]
\begin{center}
\includegraphics[width=\linewidth]{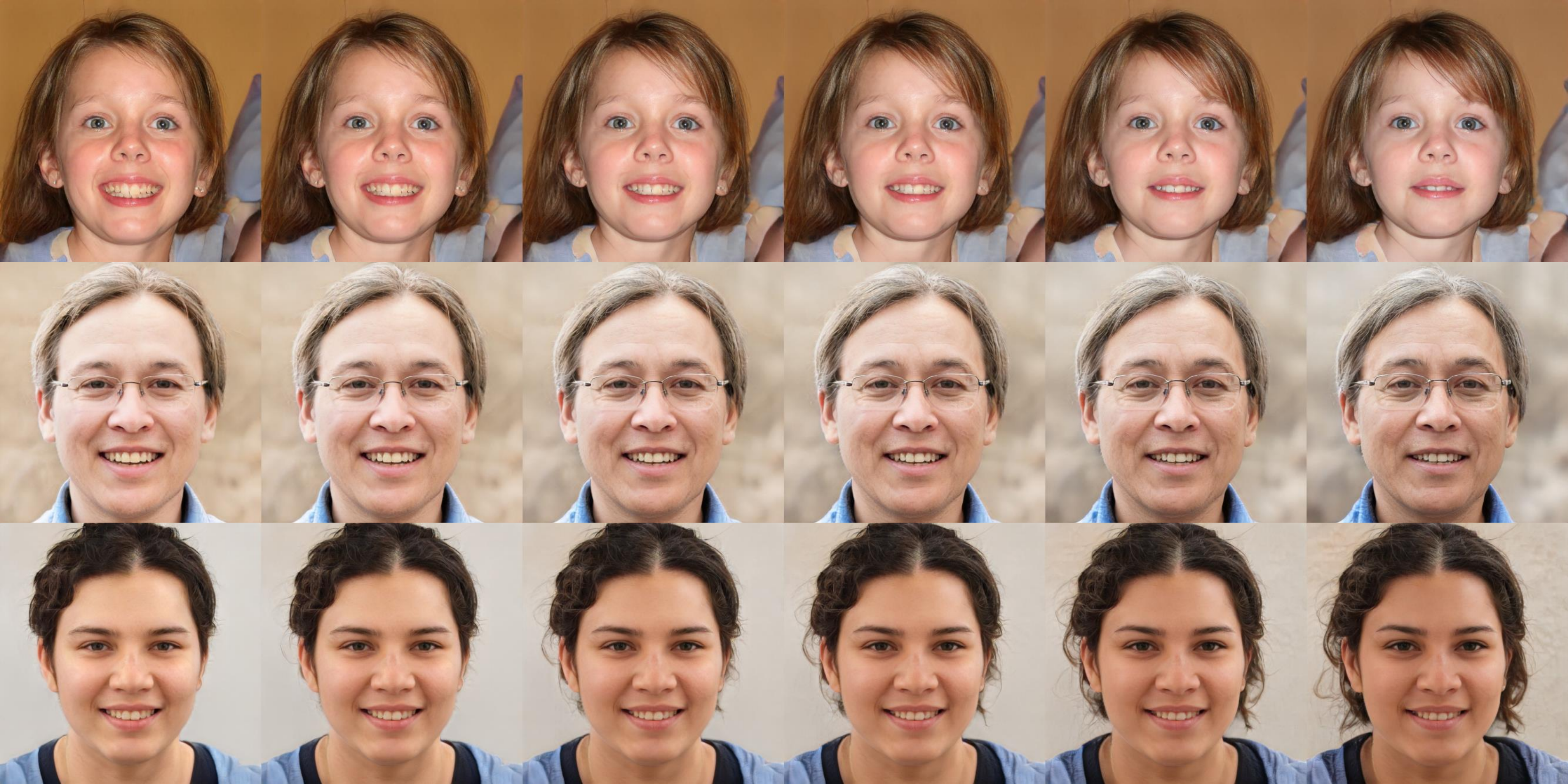}
\end{center}
   \caption{Semantic face editing using the latent projection method of \cite{shen2020interpreting}. We modify (from top to bottom): smile, age and hair length.}
\label{fig:face_editing}
\end{figure*}

In Figure~\ref{fig:linear_1024}, we encode pairs of images into the latent spaces of a SWAGAN-Bi generator and StyleGAN2's original generator using the projector provided in their original implementation. We perform linear interpolation between each pair of images and compare the results of both models. As can be seen, both latent spaces seem to behave comparably well, although our model generates sharp images throughout the interpolation process, without any of the blurring often observed during the transition (see for example the hair in the the central images of the figure). In Figure~\ref{fig:face_editing}, we show the results of semantic editing in our generator's latent space. As in the case of linear interpolation, our results do not exhibit blur around high-frequency areas such as the hair.

\begin{figure}[t]
\begin{center}
\includegraphics[width=\linewidth]{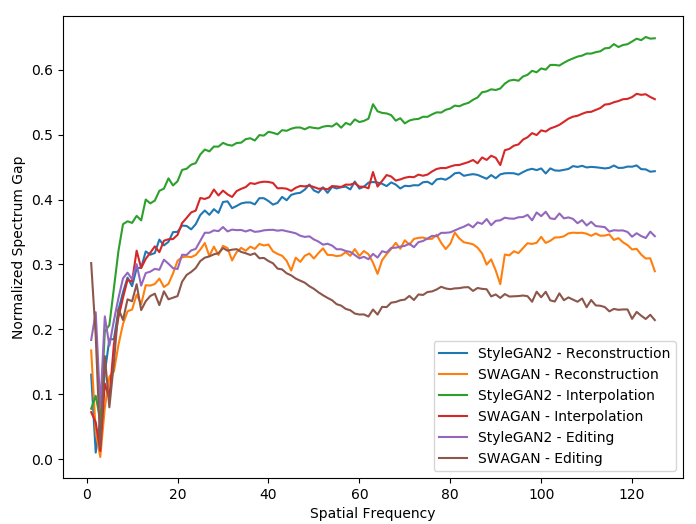}
\end{center}
   \caption{The power spectrum gap between real images and those generated by our downstream experiments. Though performance varies greatly between tasks, our high-frequency advantage is maintained for all of them.}
\label{fig:spectrum_encoded}
\end{figure}

We further analyzed the spectrum gap of images generated in all our downstream experiments. For the reconstruction experiment, we compare the spectra of $500$ reconstructed images to those of their corresponding projection targets. For the editing and interpolation experiments, we sampled random images and compared their spectra to 5000 real images from the FFHQ dataset. The results are shown in Figure~\ref{fig:spectrum_encoded}. In all scenarios, our model is able to generate more realistic spectra, indicating that our improvements can be transmitted to downstream tasks.

Our experiments show that our frequency based model preserves the versatility of StyleGAN2's latent space, allowing us to support similar operations while generating sharper results.

\begin{figure*}[t]
\begin{center}
\includegraphics[width=\linewidth]{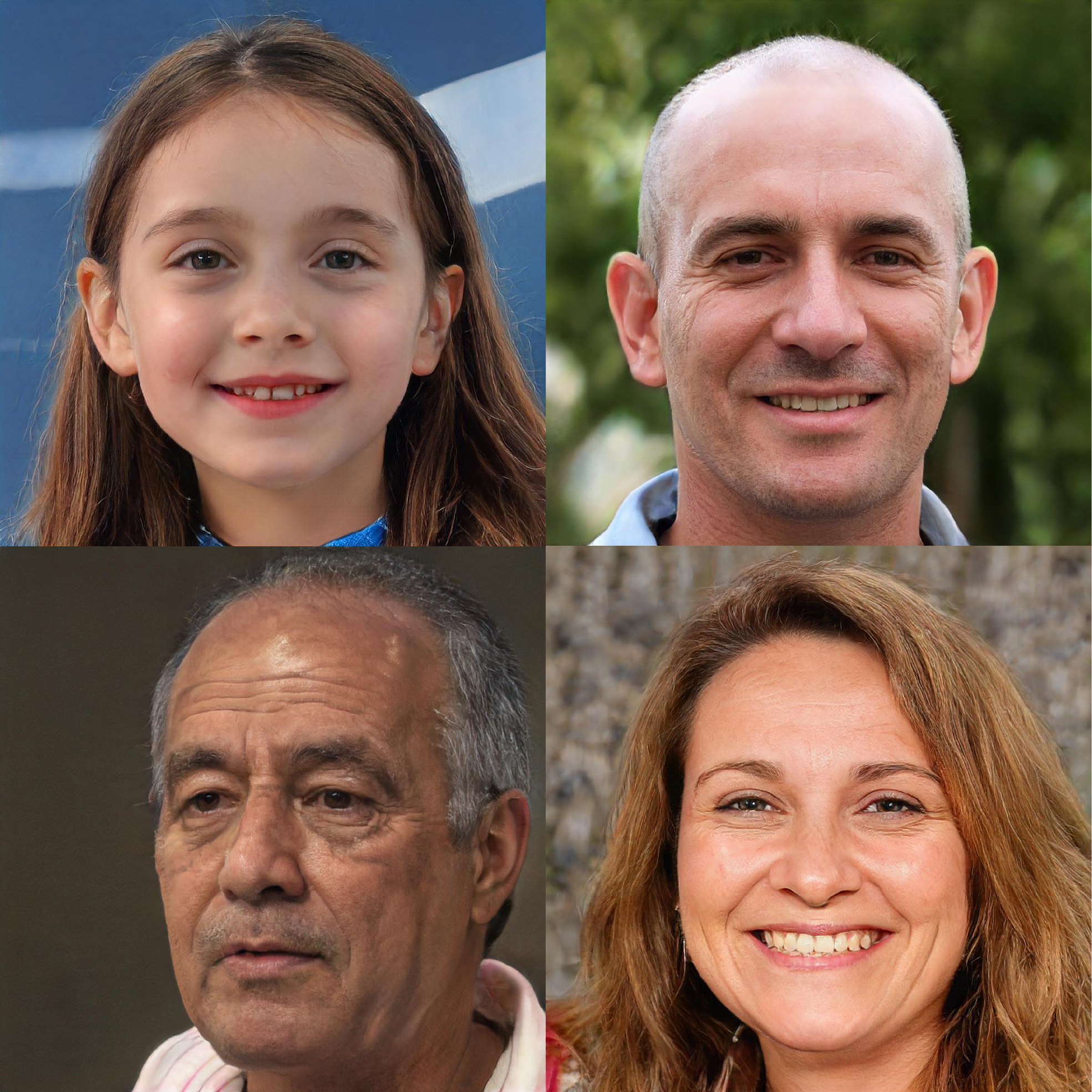}
\end{center}
   \caption{Curated face samples generated by our SWAGAN-Bi model without truncation.}
\label{fig:curated_gen_1024}
\end{figure*}

%% file: CVPR2021/future.tex
\section{Conclusions}
\label{sec:future}

We have outlined an alternative approach to the style-based generative framework, operating on the frequency domain rather than directly in RGB-space. This approach was shown to lead to more realistic visual results, particularly in the middle-to-high frequency range. By operating directly in frequency space, the model can affect high frequency changes in the output through low frequency changes in the representation, i.e. the wavelet coefficients. This  directly tackles the spectral bias of neural networks and prompts a significant increase in rate of convergence. As it turns out, adopting the wavelet-based generative approach allows for training a style-based generative model with similar performance at roughly a fourth of the computational budget.
We have further shown that our quality benefits can be extended to downstream tasks, allowing for more realistic image reconstructions or editing operations. 

The landscape of frequency-domain generation leaves many venues for future research. 
Our exploration focused on a single wavelet function (the Haar Wavelet) which may not necessarily be optimal. Indeed, it may be possible to learn a better wavelet, or to utilize a set of different wavelets, while allowing the network to mix and match between them using appropriate weights.
Another \mbox{interesting} direction for further research may be an adaptation of frequency-domain generation to the realm of videos, for example by building upon existing spatio-temporal wavelet schemes.

Where downstream tasks are concerned, we have limited ourselves to those that are tackled through direct optimization methods. Recently, however, encoder-based methods \cite{richardson2020encoding} have been gaining popularity, 
offering a faster alternative to optimization, with increasing quality for various image-to-image translation tasks. 
These encoders suffer from acute high-frequency shortcomings, in part due to their use of L2 based losses. It may therefore be beneficial to explore a similar frequency-based treatment for such encoders. Additionally, some disentanglement tasks probably lend themselves better to a spectral representation, especially those that are frequency related. This avenue is also one that should probably be explored.

In summary, we hope our work inspires others to consider alternative representation for content generation, since these can be, as demonstrated here, a more natural fit for the network to learn, yielding broad advantages in quality, runtime, and training convergence.